\documentclass[10pt,twocolumn,letterpaper]{article}

\usepackage{cvpr}              

\definecolor{cvprblue}{rgb}{0.21,0.49,0.74}
\usepackage[pagebackref,breaklinks,colorlinks,allcolors=cvprblue]{hyperref}

\usepackage{booktabs}
\usepackage{multirow}
\usepackage{graphicx}

\def\paperID{15144} 
\def\confName{CVPR}
\def\confYear{2025}

\title{DART: Disease-aware Image-Text Alignment and Self-correcting Re-alignment for Trustworthy Radiology Report Generation}

\author{Sang-Jun Park\footnotemark[1] \quad Keun-Soo Heo\footnotemark[1]  \quad Dong-Hee Shin  \quad Young-Han Son  \quad Ji-Hye Oh  \quad Tae-Eui Kam\footnotemark[2] \\
Department of Artificial Intelligence, Korea University\\
\tt\small \{wedm2401, gjrmstn1440, dongheeshin, yhson135, meeeo\_, kamte\}@korea.ac.kr
}

\begin{document}
\maketitle

{
    \renewcommand{\thefootnote}%
        {\fnsymbol{footnote}}
        \footnotetext[1]{Equal contribution.}
}
{
    \renewcommand{\thefootnote}%
        {\fnsymbol{footnote}}
        \footnotetext[2]{Corresponding author.}
}

\begin{abstract}
The automatic generation of radiology reports has emerged as a promising solution to reduce a time-consuming task and accurately capture critical disease-relevant findings in X-ray images. Previous approaches for radiology report generation have shown impressive performance. However, there remains significant potential to improve accuracy by ensuring that retrieved reports contain disease-relevant findings similar to those in the X-ray images and by refining generated reports. In this study, we propose a Disease-aware image-text Alignment and self-correcting Re-alignment for Trustworthy radiology report generation (DART) framework. In the first stage, we generate initial reports based on image-to-text retrieval with disease-matching, embedding both images and texts in a shared embedding space through contrastive learning. This approach ensures the retrieval of reports with similar disease-relevant findings that closely align with the input X-ray images. In the second stage, we further enhance the initial reports by introducing a self-correction module that re-aligns them with the X-ray images. Our proposed framework achieves state-of-the-art results on two widely used benchmarks, surpassing previous approaches in both report generation and clinical efficacy metrics, thereby enhancing the trustworthiness of radiology reports.
\end{abstract}    
\section{Introduction}
\label{sec:intro}
Radiology reports play a critical role in patient care by interpreting complex X-ray images into clear medical findings that guide diagnosis and treatment decisions. Hand-crafting radiology reports is a time-consuming task for radiologists and requires medical expertise to accurately interpret X-ray images and document findings in a clinically coherent manner \cite{liu2021contrastive, li2023auxiliary}. Consequently, the automatic generation of radiology reports has gained increasing attention in recent years as a promising solution to alleviate the task of radiologists~\cite{liu2022competence, li2023dynamic}. However, radiology report generation is inherently challenging, as it involves capturing critical medical findings, particularly disease-relevant findings, to accurately describe key descriptions in X-ray images.

Previous approaches to radiology report generation~\cite{tanida2023interactive, li2023dynamic, jing2017automatic, xue2019improved, xue2018multimodal, yang2023radiology, yin2019automatic, liu2021exploring, wang2022medical, you2021aligntransformer} have utilized encoder-decoder models inspired by image captioning methods~\cite{xu2015show, vinyals2015show, hossain2019comprehensive, pan2020x, cornia2020meshed}. Since there are significant parallels between the two tasks—both involving the generation of descriptive text from visual data—these approaches commonly use an image encoder, such as Convolutional Neural Network (CNN)~\cite{he2016deep} or Vision Transformer (ViT)~\cite{alexey2020image}, to encode X-ray images and a text decoder, such as Recurrent Neural Network (RNN)~\cite{hochreiter1997long} or Transformer~\cite{vaswani2017attention}. To enrich report generation, previous approaches incorporate prior medical knowledge, such as disease tags, entity graphs, and retrieved reports~\cite{zhang2020radiology, liu2021exploring, wang2022medical}. 

Previous approaches have demonstrated impressive performance in radiology report generation, yet there remains significant potential for improvement in two key areas: ensuring the trustworthiness of retrieved reports and refining generated outputs. First, many previous approaches incorporate retrieved reports as prior medical knowledge, often by selecting similar reports from data sources like public datasets~\cite{liu2021exploring, li2023dynamic}. However, it remains challenging to ensure that these retrieved reports contain disease-relevant findings that are closely aligned with those in the input X-ray images. For trustworthy report generation, it is essential to align disease-relevant findings in the retrieved reports with those observed in the X-ray images. Second, self-correction mechanisms have recently emerged as an effective strategy for improving the quality of generated texts by reducing errors through self-feedback~\cite{welleck2022generating, madaan2024self}. Beyond previous approaches, self-correction mechanisms hold great potential for enhancing radiology report generation, especially in capturing disease-relevant findings in X-ray images.

In this study, we propose a Disease-aware image-text Alignment and self-correcting Re-alignment for Trustworthy radiology report generation (DART), a novel framework that ensures retrieved reports contain similar disease-relevant findings and introduces a self-correction mechanism to refine generated reports. Firstly, we generate initial reports based on image-to-text retrieval with disease-matching, which retrieves reports containing disease-relevant findings similar to those in the input images. We embed both images and reports into a shared embedding space using contrastive learning with a disease-matching constraint, ensuring the retrieval of reports containing disease-relevant findings that closely align with the input images. Additionally, we construct a disease classifier to extract disease-relevant features, which guide the report generation process to reflect the disease-relevant findings in the input images. Secondly, we introduce a self-correction module designed to refine the initial reports by re-aligning them with the input image features within the embedding space. After reports are generated in the first stage, we re-align the initial reports with their corresponding images in the shared embedding space. To refine the generated reports, the self-correction module is trained to align the generated reports more closely with the images.

Our contributions can be summarized as follows:
\begin{itemize}[noitemsep, topsep=0pt]
\item To the best of our knowledge, our proposed framework is the first to introduce a self-correction mechanism for radiology report generation by re-aligning an image-text embedding space, advancing beyond previous approaches.

\item We propose a trustworthy report generation model by disease-aware image-text alignment, which ensures capturing critical disease-relevant findings in X-ray images.

\item Our proposed framework demonstrates promising performance on two widely used benchmarks, outperforming state-of-the-art methods in radiology report generation and clinical efficacy metrics.
\end{itemize}
\begin{figure*}[t]
	\begin{center}
		\includegraphics[width=\linewidth]{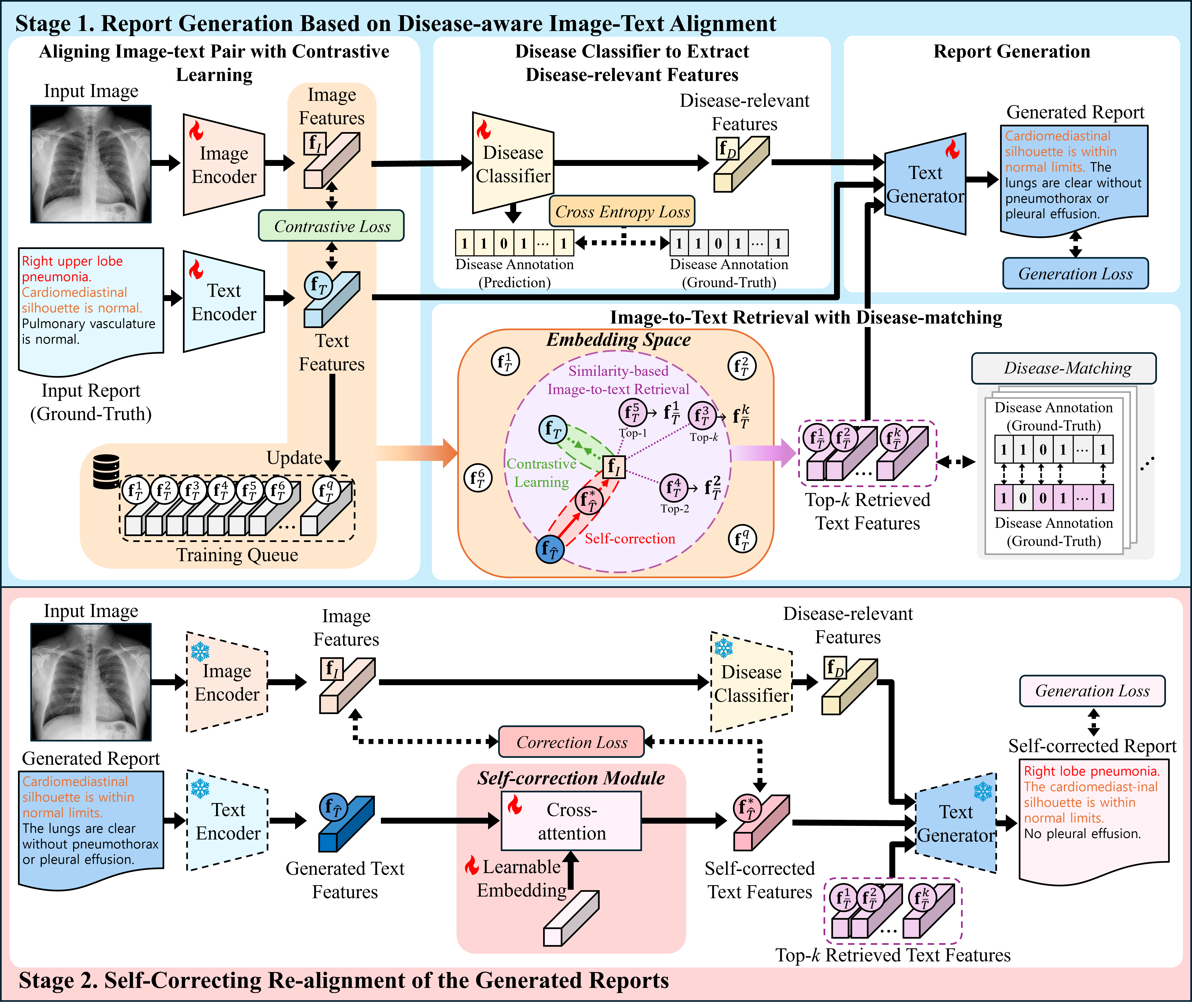}
	\end{center}
	\caption{An overview of our proposed framework, which consists of two stages: (1) report generation based on disease-aware image-text alignment and (2) self-correcting re-alignment of generated reports. In the first stage, our proposed framework generates initial reports by text features, disease-relevant features, and retrieved text features that are closely aligned with image features in an embedding space. In the second stage, a self-correction mechanism refines the generated reports by re-aligning them within the embedding space to further enhance consistency with the input images.}
	\label{fig:method_overview}
\end{figure*}
\section{Related Work}
\label{sec:related}

\textbf{Image Captioning.}
Image captioning aims to generate a descriptive sentence for a given image, and it has been extensively studied in computer vision. Most approaches~\cite{anderson2018bottom,cornia2020meshed,pan2020x,rennie2017self,vinyals2015show,xu2015show,yang2016review} typically follow an encoder-decoder framework, where the image encoder (e.g., CNN or ViT) is used to encode image features, and the decoder (e.g., RNN or Transformer) is used to generate text~\cite{liu2022competence,wang2023metransformer}. However, these approaches are challenging for accurate generation of radiology reports due to two key aspects. First, X-ray images contain both normal and abnormal regions, requiring capturing detailed disease-relevant findings. Second, radiology reports are longer than typical image captions, consisting of multiple sentences that describe both the normal findings and any abnormalities. As a result, simply applying conventional image captioning techniques to radiology reports leads to a dominance of normal findings~\cite{liu2022competence,li2023dynamic,tanida2023interactive}. The failure to accurately capture abnormal findings remains a well-known limitation in this domain~\cite{liu2022competence,tanida2023interactive,li2023dynamic}.

\noindent\textbf{Medical Report Generation.}
Most studies on radiology report generation can generally be divided into two primary approaches. 
The first approach focuses on improving the encoder-decoder architecture, and it also emphasize aligning visual and textual information to generate more consistent reports. For example, many studies~\cite{jing2017automatic, xue2019improved, xue2018multimodal, yang2023radiology, yin2019automatic} employ LSTM networks with hierarchical structures to effectively manage the descriptive characteristics of radiology reports.
Tanida et al.~\cite{tanida2023interactive} developed an image encoder that enhances visual features by focusing on anatomical regions within X-ray images. Li et al.~\cite{li2023dynamic} proposed a novel framework employing contrastive learning paradigms for radiology reporting, utilizing a dynamic graph to enhance visual representations. Wang et al.~\cite{wang2023metransformer} introduced multiple expert tokens into the transformer encoder and decoder; in the encoder, these tokens focus on different image regions, while in the decoder, they guide the interaction between input words and visual tokens to generate the reports. Liu et al.~\cite{liu2024context} employ a multi-modal contextual vector to effectively capture and represent the contextual details, enhancing the understanding of both visual and textual information within the model. Liu et al.~\cite{liu2024bootstrapping} propose an approach to tailor LLMs for the medical domain and enhance the quality of report generation. Lastly, Shen et al.~\cite{shen2024automatic} propose an approach that queries a memory matrix based on a combination of visual features and positional embeddings.
The second approach focuses on utilizing medical knowledge to inform the report generation process.
Some studies~\cite{wang2022medical, you2021aligntransformer} incorporate disease tags that relate directly to the patient's medical conditions. Zhang et al.~\cite{zhang2020radiology} and Liu et al.~\cite{liu2021exploring} utilized a universal graph of 20 entities with connections representing the relationships between entities related to the same organ. Additionally, Liu et al.~\cite{liu2021exploring} incorporated global representations from pre-retrieved reports in the training corpus to represent domain-specific knowledge. 
Li et al.~\cite{li2023dynamic} dynamically updated the graph by injecting new knowledge extracted from reports for each case, rather than using a fixed graph as in~\cite{zhang2020radiology}. Finally, Hou et al.~\cite{hou2024energy} directly utilized diverse off-the-shelf medical expert models or knowledge to design energy function. 

\noindent\textbf{Self-correction.}
Self-correction has emerged as a solution to improve the quality of generated outputs through refinement~\cite{welleck2022generating, madaan2024self}. This concept has been applied in areas such as natural language processing~\cite{welleck2022generating,madaan2024self, yan2024corrective}, where models benefit from feedback to reduce errors and enhance accuracy over time~\cite{welleck2022generating}. This feedback enables to improve output accuracy, coherence, and alignment with task-specific constraints. To the best of our knowledge, our proposed framework is the first approach to apply self-correction to the radiology report generation task, where accuracy and consistency with X-ray images are crucial. By implementing a self-correction mechanism, we refine initial reports by aligning them closely with image features of X-ray images. 

\newpage
\section{Method}
Our proposed framework comprises two stages: (1) report generation based on disease-aware image-text alignment and (2) self-correcting re-alignment of the generated reports, as shown in Fig. \ref{fig:method_overview}. In the first stage, we introduce a disease-aware report generation model to generate reports from input images by leveraging image-to-text retrieval with a disease-matching constraint, ensuring that the generated reports accurately reflect disease-relevant findings. Additionally, we enhance the report generation model by extracting disease-relevant features through disease classifier. In the second stage, we propose a self-correction module to refine generated reports by further aligning them with the image features in the embedding space, enhancing both disease classification and report generation.

\subsection{Report Generation Based on Disease-aware Image-Text Alignment}
We present a trustworthy report generation model based on disease-aware image-to-text alignment. First, an input image and its associated report are embedded into a shared embedding space. Next, disease-relevant features are extracted through disease classifier. Then, reports with similar disease-relevant findings are retrieved using image-to-text retrieval with disease-matching constraints. Finally, accurate reports are generated based on the retrieved reports and disease-relevant features, ensuring that the generated report captures essential disease-relevant findings.

\noindent\textbf{Aligning Image-Text Pair with Contrastive Learning.}
\label{sec:method_contrastive}
We embed X-ray images and their corresponding reports into an embedding space through contrastive learning, ensuring that images and reports are closely aligned. We construct two encoders: an image encoder and a text encoder. The image encoder produces image features $\mathbf{f}_I \in \mathbb{R}^{d \times e}$ from an input image $I \in \mathbb{R}^{v \times h \times w}$, while the text encoder generates text features $\mathbf{f}_T \in \mathbb{R}^{d \times e}$ from the associated report $T \in \mathbb{R}^{l \times e}$. Here, $v$ denotes the number of views of the input image, $h$ and $w$ are the height and width of the input image, respectively, $l$ denotes the length of the associated report, $e$ denotes the dimensions of the embedding space, and $d$ represents the number of ground-truth disease annotations. Following \cite{nguyen2021automated}, the ground-truth disease annotations consist of labeled disease keywords and disease-relevant keywords in the datasets. We use CLIP loss \cite{radford2021learning} as a contrastive loss $\mathcal{L}_{\text{con}}$, which aligns image and text embeddings by maximizing the cosine similarity between paired image-text embeddings, i.e., an image and its corresponding report, and minimizing the similarity between unpaired image-text embeddings. By leveraging the contrastive loss, we create the embedding space that is essential for the subsequent image-to-text retrieval step, enabling the retrieval of the most relevant textual features based on the input image features. Further details on the contrastive loss can be found in the supplementary materials.

\noindent\textbf{Disease Classifier to Extract Disease-relevant Features.}
To extract disease-relevant features, we construct a disease classifier that uses a cross-attention mechanism~\cite{vaswani2017attention} on the image features to predict the ground-truth disease annotations. The disease classifier is optimized by minimizing the classification loss to encourage accurate predictions of the disease annotations. The classification loss $\mathcal{L}_{cls}$ is defined as:

\begin{equation}
    \hat{\mathbf{y}} = \textit{Softmax}\left(\frac{\mathbf{f}_I \cdot \mathbf{\Phi}^T}{\sqrt{e}}\right),
\end{equation}
\begin{equation}
    \mathcal{L}_{cls} = \textit{Cross-entropy}(\hat{\mathbf{y}}, \mathbf{y}),
\end{equation}
where $\hat{\mathbf{y}} \in \mathbb{R}^{d \times 2}$ represents the predicted disease annotations, $\mathbf{\Phi} \in \mathbb{R}^{2 \times e}$ is a learnable embedding for the disease classifier, $\mathbf{y} \in \mathbb{R}^{d \times 2}$ is the ground-truth disease annotations, $\textit{Softmax}$ is the softmax function, and $\textit{Cross-entropy}$ denotes the cross-entropy loss \cite{mao2023cross}.

Next, we extract disease-relevant features accurately to capture disease-relevant findings in the image. The disease-relevant features, denoted as $\mathbf{f}_D \in \mathbb{R}^{d \times e}$, integrate the predicted disease annotations with the image features. The disease-relevant features $\mathbf{f}_D$ are formulated as:

\begin{equation}
    \mathbf{f}_D = \hat{\mathbf{y}} \cdot \mathbf{\Phi} + \mathbf{f}_I.
\end{equation}

\noindent\textbf{Image-to-text Retrieval with Disease-matching.}
After aligning the image and text embeddings in a shared embedding space using contrastive learning, we retrieve reports relevant to the image features. Specifically, we calculate similarity, such as cosine similarity, between the image embedding and text embeddings stored in a training queue that maintains recent text embeddings from other images. Based on these similarity scores, we retrieve the top-$k$ text features ($\mathbf{f}_{\bar{T}}^1, \mathbf{f}_{\bar{T}}^2, ..., \mathbf{f}_{\bar{T}}^k$) with the highest similarity to the image features. These retrieved text features represent reports that contain disease-relevant findings similar to those in the input image.

Additionally, to ensure that the retrieved text features contain similar disease-relevant findings, we introduce a disease-matching constraint, which assesses the difference between the ground-truth disease annotations of the input image and the retrieved texts. Formally, the disease-matching constraint $\gamma$ is defined as:

\begin{equation} \gamma = \frac{1}{k} \sum_{i=1}^{k} \textit{Cross-Entropy}(\mathbf{y}, \mathbf{y}_{\bar{T}}^i), \end{equation}
where $\mathbf{y}$ is the ground-truth disease annotation of the input image, $\mathbf{y}_{\bar{T}}^i$ represents the ground-truth disease annotation of the $i^{th}$ retrieved text features, and $k$ is the number of retrieved texts.

We minimize the disease-matching constraint to encourage the encoders to match the ground-truth disease annotations of the input image and the retrieved texts, ensuring the retrieved texts contain disease-relevant findings similar to those in the input image.

\noindent\textbf{Report Generation.} We construct a text generator to synthesize a trustworthy report that accurately describes key findings in X-ray images, using the retrieved text features, disease-relevant features, and text features. To train the text generator, we employ a generation loss that minimizes the discrepancy between the generated report and the ground-truth report. The generation loss $\mathcal{L}_{gen}$ is computed using the auto-regressive loss, i.e., the cross-entropy loss, which penalizes differences between the generated report $\hat{T}$ and the ground-truth report $T$. Further details on the generation loss can be found in the supplementary materials.

In the first stage, we minimize a total loss $\mathcal{L}_{stage1}$, which consists of the contrastive loss, the disease-matching constraint, the classification loss, the generation loss to generate a radiology report that accurately describes key findings in the input image:

\begin{equation}
\mathcal{L}_{stage1} = \mathcal{L}_{con} + \lambda_{cls} \cdot \mathcal{L}_{cls} + \lambda_{gen} \cdot \mathcal{L}_{gen} + \lambda_m \cdot \gamma,
\end{equation}
where $\lambda_{cls}$, $\lambda_{gen}$, and $\lambda_{m}$ are weighting coefficients for the classification loss, the generation loss, and the disease-matching constraint, respectively, and are set to $1$, $1$, and $10$.

\begin{table*}[t]
\centering
\begin{tabular}{c|l|cccccc}
\hline
Dataset                     & Method                                     & BLEU-1         & BLEU-2         & BLEU-3         & BLEU-4         & RG-L           & METEOR         \\ \hline
\multirow{12}{*}{MIMIC-CXR} & R2Gen~\cite{chen2020generating}            & 0.353          & 0.218          & 0.145          & 0.103          & 0.277          & 0.142          \\
                            & R2GenCMN~\cite{chen2022cross}              & 0.353          & 0.218          & 0.148          & 0.106          & 0.278          & 0.142          \\
                            & PPKED~\cite{liu2021exploring}              & 0.360          & 0.224          & 0.149          & 0.106          & 0.284          & 0.149          \\
                            & CMCL~\cite{liu2022competence}              & 0.344          & 0.217          & 0.140          & 0.097          & 0.281          & 0.133          \\
                            & DCL~\cite{li2023dynamic}                   & -              & -              & -              & 0.109          & 0.284          & 0.150          \\
                            & RGRG~\cite{tanida2023interactive}          & 0.373          & 0.249          & 0.175          & 0.126          & 0.264          & 0.168          \\
                            & METransformer~\cite{wang2023metransformer} & 0.386          & 0.250          & 0.169          & 0.124          & 0.291          & 0.152          \\
                            & ECRG~\cite{hou2024energy}                  & 0.379          & 0.253          & 0.175          & 0.123          & 0.266          & 0.164          \\
                            & Med-LLM~\cite{liu2024context}              & -              & -              & -              & 0.128          & 0.289          & 0.161          \\
                            & MA~\cite{shen2024automatic}                & 0.396          & 0.244          & 0.162          & 0.115          & 0.274          & 0.151          \\
                            & I3 + C2FD~\cite{liu2024bootstrapping}      & 0.402          & 0.262          & 0.180          & 0.128          & 0.291          & \textbf{0.175} \\ \cline{2-8} 
                            & Ours                                       & \textbf{0.437} & \textbf{0.279} & \textbf{0.191} & \textbf{0.137} & \textbf{0.310} & \textbf{0.175} \\ \hline
\multirow{11}{*}{IU X-ray}  & SentSAT+KG~\cite{zhang2020radiology}       & 0.441          & 0.291          & 0.203          & 0.147          & 0.367          & -              \\
                            & R2Gen~\cite{chen2020generating}            & 0.470          & 0.304          & 0.219          & 0.165          & 0.371          & 0.187          \\
                            & R2GenCMN~\cite{chen2022cross}              & 0.475          & 0.309          & 0.222          & 0.170          & 0.375          & 0.191          \\
                            & PPKED~\cite{liu2021exploring}              & 0.483          & 0.315          & 0.224          & 0.168          & 0.376          & 0.190          \\
                            & CMCL~\cite{liu2022competence}              & 0.473          & 0.305          & 0.217          & 0.162          & 0.378          & 0.186          \\
                            & MSAT~\cite{wang2022medical}                & 0.481          & 0.316          & 0.226          & 0.171          & 0.372          & 0.190          \\
                            & METransformer~\cite{wang2023metransformer} & 0.483          & 0.322          & 0.228          & 0.172          & 0.380          & 0.192          \\
                            & Med-LLM~\cite{liu2024context}              & -              & -              & -              & 0.168          & 0.381          & 0.209          \\
                            & MA~\cite{shen2024automatic}                & \textbf{0.501} & 0.328          & 0.230          & 0.170          & 0.386          & \textbf{0.213} \\
                            & I3 + C2FD~\cite{liu2024bootstrapping}      & 0.499          & 0.323          & 0.238          & 0.184          & 0.390          & 0.208          \\ \cline{2-8} 
                            & Ours                                       & 0.486          & \textbf{0.348} & \textbf{0.265} & \textbf{0.208} & \textbf{0.411} & 0.205          \\ \hline
\end{tabular}%
\caption{A comparison of descriptive accuracy between our proposed framework (Ours) and state-of-the-art methods using BLEU scores (BLEU-1 to BLEU-4), ROUGE-L (RG-L), and METEOR on the MIMIC-CXR (upper section) and IU X-ray (lower section) datasets.}
\label{tab:result}
\end{table*}
\begin{table}[]
\centering
\begin{tabular}{l|ccc}
\hline
Model                                      & F1    & Precision      & Recall \\ \hline
R2Gen~\cite{chen2020generating}            & 0.276 & 0.333          & 0.273  \\
R2GenCMN~\cite{chen2022cross}              & 0.278 & 0.334          & 0.275  \\
METransformer~\cite{wang2023metransformer} & 0.311 & 0.364          & 0.309  \\
Med-LLM~\cite{liu2024context}                                   & 0.395 & 0.412 & 0.373  \\
MA~\cite{shen2024automatic}                                        & 0.389 & 0.411          & 0.398  \\ 
RGRG~\cite{tanida2023interactive}      & 0.447 & 0.461     & 0.475  \\
I3 + C2FD~\cite{liu2024bootstrapping} & 0.473 & 0.465     & 0.482 \\ \hline
Ours & \textbf{0.533} & \textbf{0.520} & \textbf{0.546}\\
(Disease Classification)  & (0.427) & (0.404) & (0.506)  \\ \hline
\end{tabular}%
\caption{A comparison of the clinical efficacy (CE) metrics between our proposed framework (Ours) and state-of-the-art methods using F1 score, precision, and recall on the MIMIC-CXR dataset. Also, we evaluate the disease classifier performance.}
\label{tab:classification}
\end{table}

\begin{table*}[t]
\centering
\begin{tabular}{c|c|cccc|cccccc}
\hline
Dataset                    & Setting & CL         & I2T        & DM         & SC         & BLEU-1         & BLEU-2         & BLEU-3         & BLEU-4         & RG-L           & METEOR         \\ \hline
\multirow{5}{*}{MIMIC-CXR} & BASE    &     -       &      -      &       -     &      -      & 0.371          & 0.233          & 0.157          & 0.111          & 0.277          & 0.153          \\
                           & (a)     & \checkmark &     -       &       -     &       -     & 0.383          & 0.241          & 0.162          & 0.113          & 0.279          & 0.156          \\
                           & (b)     & \checkmark & \checkmark &     -       &        -    & 0.400          & 0.254          & 0.172          & 0.121          & 0.303          & 0.164          \\
                           & (c)     & \checkmark & \checkmark & \checkmark &       -     & 0.418          & 0.266          & 0.181          & 0.129          & 0.308          & 0.169          \\ \cline{2-12} 
                           & (d)     & \checkmark & \checkmark & \checkmark & \checkmark & \textbf{0.437} & \textbf{0.279} & \textbf{0.191} & \textbf{0.137} & \textbf{0.310} & \textbf{0.175} \\ \hline
\end{tabular}
\caption{An ablation study of our proposed framework on the MIMIC-CXR dataset, assessing the impact of key components: contrastive loss (CL), image-to-text retrieval (I2T), disease-matching constraint (DM), and self-correction (SC). A ``\checkmark" indicates the presence of each component, while ``-" denotes its absence. The BASE setting involves training only the classifier and generator.}
\label{tab:ablation}
\end{table*}

\subsection{Self-Correcting Re-alignment of Generated Reports}
To further refine the report generated in stage 1, we introduce a self-correction mechanism that re-aligns the generated report with its corresponding image features in the shared embedding space. When the generated report is embedded, a subtle gap may exist between the report and the image in the embedding space. To address this, we minimize the gap by re-aligning the generated report, effectively refining it to better capture critical findings in the images.

We first embed the generated report $\hat{T}$ into the embedding space using the text encoder trained in stage 1, obtaining the generated text features $\mathbf{f}_{\hat{T}} \in \mathbb{R}^{d \times e}$. The generated text features are processed by a self-correction module, which refines their alignment with the corresponding image features. The self-correction module, equipped with a learnable embedding $\Psi$, applies a cross-attention mechanism to extract self-corrected text features $\mathbf{f}_{\hat{T}}^* \in \mathbb{R}^{d \times e}$ by minimizing the distance between them, resulting in self-corrected text features $\mathbf{f}_{\hat{T}}^* \in \mathbb{R}^{d \times e}$. The self-correction module can be expressed as:

\begin{equation} \mathbf{f}_{\hat{T}}^* = \textit{Softmax}\left(\frac{\mathbf{f}_{\hat{T}} \cdot \mathbf{\Psi}^T}{\sqrt{e}}\right) \mathbf{\Psi},
\end{equation}
where $\mathbf{\Psi} \in \mathbb{R}^{2 \times e}$ represents the learnable embedding for self-correction, and $\mathbf{f}_{\hat{T}}$ is the generated text features from stage 1.

To optimize the self-correction module, we introduce a correction loss that measures the similarity, specifically cosine similarity, between the self-corrected text features and the image features, encouraging the model to minimize any errors or omissions in the generated report. The correction loss $\mathcal{L}_{cor}$ is defined as:
\begin{equation} \mathcal{L}_{cor} = 1 - \frac{\mathbf{f}_{\hat{T}}^* \cdot \mathbf{f}_I}{|\mathbf{f}_{\hat{T}}^*| \cdot |\mathbf{f}_I|}.
\end{equation}

By minimizing the correction loss, the self-correction module learns to align the generated text features with the image features, ensuring that the generated report is refined semantically to maintain consistency with the image features.

Finally, we generate a self-corrected report by passing the self-corrected text features, top-$k$ retrieved texts, and disease-relevant features through the text generator. The self-correction module is optimized by minimizing the total loss $\mathcal{L}_{stage2}$, which consists of generation loss and correction loss. The total loss $\mathcal{L}_{stage2}$ is defined as:
\begin{equation}
\mathcal{L}_{stage2} = \mathcal{L}_{gen} + \lambda_{cor} \cdot \mathcal{L}_{cor},
\end{equation}
where $\lambda_{cor}$ is a weighting coefficient that adjusts the correction loss and is set to $5$, and $\mathcal{L}_{gen}$ is the generation loss, i.e., auto-regressive loss. In stage 2, only the self-correction module is trained, while the other modules remain frozen.

\section{Experiments}
\subsection{Datasets}

\noindent\textbf{MIMIC-CXR.} MIMIC-CXR~\cite{johnson2019mimic} dataset is the most extensive publicly available dataset, containing $227,835$ radiology reports from patients examined at the Beth Israel Deaconess Medical Center. In this study, we adopt the official split of the MIMIC-CXR dataset to ensure a fair evaluation comparison, consistent with previous studies~\cite{chen2020generating,wang2023metransformer}.

\noindent\textbf{IU X-ray.} Indiana University Chest X-ray Collection~\cite{demner2016preparing} (IU X-ray) includes $7,470$ chest X-ray images and $3,955$ corresponding reports. Each report is linked to either frontal images alone or a combination of frontal and lateral view images. Following~\cite{chen2020generating,wang2023metransformer}, we divide the dataset into training, testing, and validation sets in a ratio of $7$:$1$:$2$.

\subsection{Experimental Settings}
\noindent\textbf{Evaluation Metrics.} 
Following~\cite{chen2020generating,chen2022cross,wang2023metransformer}, we utilize CheXpert~\cite{irvin2019chexpert} to label the generated reports and assess clinical efficacy metrics using F1, Precision, and Recall Scores. We assess the descriptive accuracy of the generated reports by employing commonly used natural language generation (NLG) metrics, such as BLEU-1 to BLEU-4~\cite{papineni2002bleu}, METEOR~\cite{banerjee2005meteor}, and ROUGE-L~\cite{lin2004rouge}. 


\noindent\textbf{Implementation Details.} 
For the image encoder, we utilize ResNet-50~\cite{he2016deep} pre-trained on ImageNet~\cite{deng2009imagenet}. For the text encoder and decoder, we utilize a Transformer~\cite{vaswani2017attention} encoder and a Transformer decoder, respectively. Transformer has 8 heads and dimension of $e$ = $256$. We use the AdamW~\cite{loshchilov2017decoupled} optimizer with a batch size of $8$, a learning rate of $3$e-$4$, and a weight decay of $0.01$. We set $k$ to $3$ in top-$k$ retrieval.

\subsection{Comparison with State-of-the-art Methods}

\noindent\textbf{Descriptive Accuracy.} We compare the performance of our proposed framework with several state-of-the-art methods on two widely used benchmarks: MIMIC-CXR and IU X-ray. Table~\ref{tab:result} demonstrates the effectiveness of our proposed framework in generating accurate and descriptive radiology reports. On the MIMIC-CXR dataset, our proposed framework achieves the highest scores across all metrics, outperforming state-of-the-art methods in BLEU-1 (0.437), BLEU-2 (0.279), BLEU-3 (0.191), BLEU-4 (0.137), ROUGE-L (0.310), and METEOR (0.175). On the IU X-ray dataset, our proposed framework demonstrates similarly strong performance, achieving the highest scores in BLEU-2 (0.348), BLEU-3 (0.265), BLEU-4 (0.208), and ROUGE-L (0.411). While MA~\cite{shen2024automatic} achieves the best BLEU-1 and METEOR scores, our proposed framework excels in generating longer, contextually relevant reports with high BLEU-2 to BLEU-4 scores and the highest ROUGE-L.

These results highlight the trustworthiness of our proposed framework, particularly the benefits of incorporating disease-aware report generation with self-correction. Our proposed framework not only captures essential medical findings but also maintains coherence, enhancing both the accuracy and clinical relevance of the generated reports.

\begin{figure*}[]
	\begin{center}
		\includegraphics[width=\linewidth]{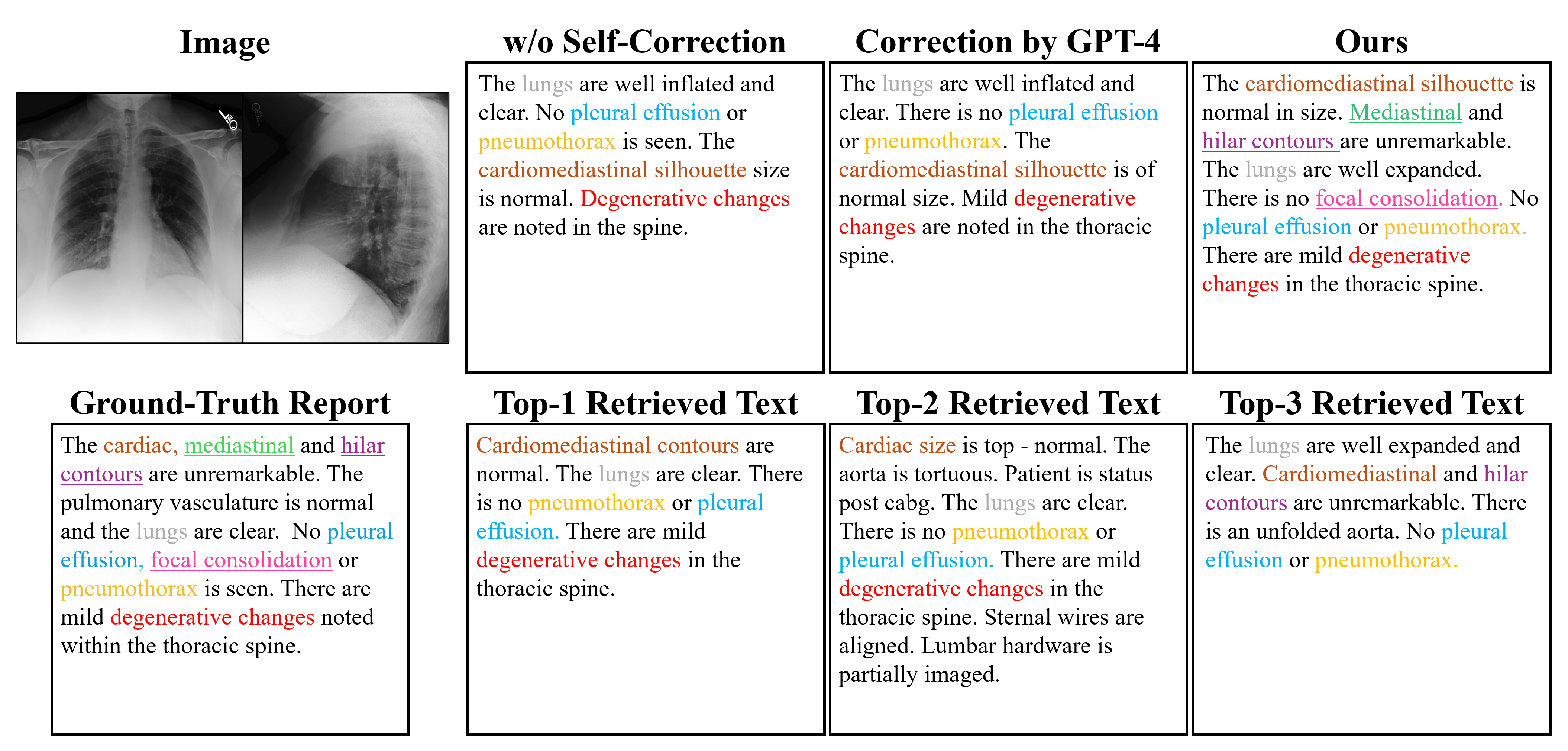}
	\end{center}
	\caption{A qualitative analysis of reports for a sample from the MIMIC-CXR dataset is presented. The top row displays an image set from two different views alongside a generated report from our proposed framework without the self-correction module (``w/o Self-Correction”). We further attempt to refine the generated report of ``w/o Self-Correction”  using GPT-4 \cite{achiam2023gpt} (``Correction by GPT-4") to compare it with the generated report from our proposed framework with self-correction (``Ours”). The bottom row shows the ground-truth report and the top-$3$ retrieved texts from image-to-text retrieval. Key findings are highlighted in different colors for clarity.}
	\label{fig:discussion}
\end{figure*}

\noindent\textbf{Clinical Efficacy Metrics \& Disease Classification.} Table \ref{tab:classification} presents a comparison of the clinical efficacy (CE) metrics between our proposed framework and state-of-the-art methods on the MIMIC-CXR dataset, evaluated by F1 score, precision, and recall. Our proposed framework significantly outperforms state-of-the-art methods, showing the highest F1 score, precision, and recall. Additionally, we evaluate the disease classifier performance, achieving strong performance. These results highlight that the generated reports of our proposed framework effectively capture disease-relevant findings while confirming that the disease classifier accurately extracts critical disease-related features.

\begin{figure*}[]
	\begin{center}
		\includegraphics[width=\linewidth]{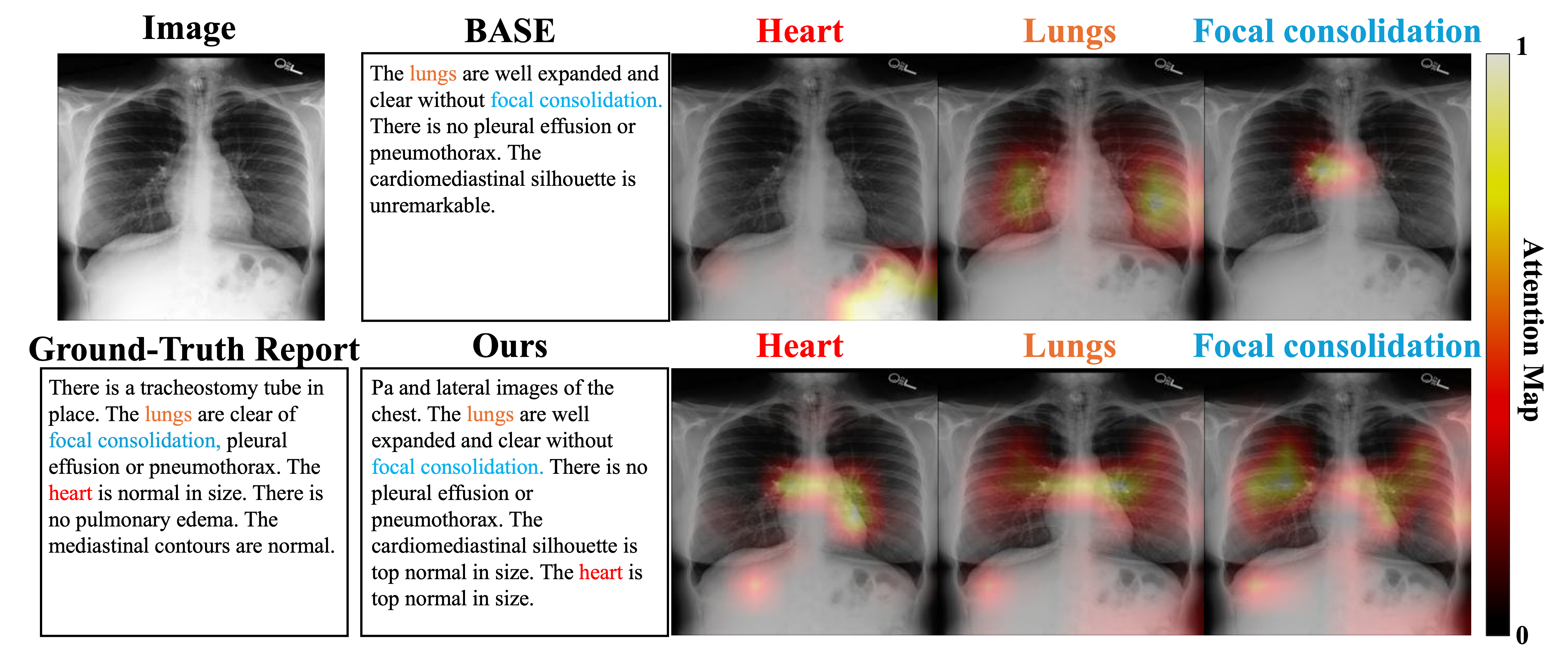}
	\end{center}
	\caption{A visualization of the generated reports and attention maps from the baseline model (BASE) and our proposed framework (Ours) on one sample from the MIMIC-CXR dataset. The attention maps, visualized using Grad-CAM \cite{selvaraju2017grad}, illustrate the regions that BASE and Ours focuses on according to three keywords ``heart,” ``lung,” and ``focal consolidation,” with each keyword highlighted in a different color.}
	\label{fig:visualization}
\end{figure*}

\subsection{Ablation Study}

We present an ablation study to evaluate the incremental effect of each key component in our proposed framework: contrastive loss (CL), image-to-text retrieval (I2T), disease-matching constraint (DM), and self-correction (SC). Table \ref{tab:ablation} shows performance improvements on the MIMIC-CXR dataset as these components are progressively added to the BASE setting with only the disease classifier and generator.

Starting from the BASE setting, adding CL in setting (a) provides a modest improvement across all metrics, as CL aligns image and text embeddings more effectively, enabling better feature representations. Incorporating I2T in setting (b) boosts BLEU and ROUGE-L scores, highlighting the positive impact of retrieving disease-relevant reports based on image features. The addition of the (DM) in setting (c) results in further gains by ensuring that retrieved text features are more aligned with the disease-relevant findings of the input images.

Finally, incorporating SC in setting (d) yields the highest performance, achieving significant improvements across all evaluation metrics. This highlights that the self-correction module effectively refines the generated reports by re-aligning them with the input image features in the embedding space. 


\section{Discussion}
\noindent\textbf{Qualitative Analysis.}
Fig. \ref{fig:discussion} presents a qualitative analysis of generated reports from the MIMIC-CXR dataset, including generated report from our proposed framework without self-correction (``w/o Self-Correction''). And we refine the generated report of ``w/o Self-Correction" by GPT-4 \cite{achiam2023gpt} (``Correction by GPT-4''), and our proposed framework with self-correction (``Ours''). We also show the ground-truth report and the top-3 retrieved reports from image-to-text retrieval. Details are provided in the supplementary material.

``Ours'' captures more key findings of the ground-truth report compared to both ``w/o Self-Correction'' and ``Correction by GPT-4''. In detail, ``w/o Self-Correction'' captures most key findings of the ground-truth report, but it omits some findings such as focal consolidation, and while GPT-4 refinement improves phrasing, it fails to address these omissions, resulting in a similar report.
The top-3 retrieved reports provide additional context and contain key findings aligned with the ground-truth report, such as cardiomediastinal silhouette and degenerative changes. This demonstrates that our proposed framework effectively leverages the retrieved reports similar to the ground-truth. 

In summary, this analysis highlights that self-correction in our framework is effective in capturing key findings in X-ray images by re-aligning the generated reports in the embedding space.  Additionally, our proposed approach retrieves reports with disease-relevant findings that closely align with those in the input X-ray images.

\noindent\textbf{Attention Visualization.}
Fig.~\ref{fig:visualization} presents attention visualizations using Grad-CAM~\cite{selvaraju2017grad} to compare our proposed framework (“Ours”) with the ``BASE'' setting from the ablation study. The visualization highlights the regions of focus for crucial disease-related findings, such as ``heart,'' ``lungs,'' and ``focal consolidation,'' with different colors.

``BASE'' predominantly attends to regions associated with ``lungs'' but fails to focus on key areas related to the ``heart.'' This lack of precise focus is reflected in its generated report. On the other hand, ``Ours'' demonstrates precise attention, correctly identifying the heart region and aligning well with the ground-truth report. Similarly, the attention maps for ``lungs'' and ``focal consolidation'' are more consistent with the corresponding areas in the image, leading to a more accurate and clinically relevant generated report. This demonstrates that our proposed framework effectively captures disease-relevant features, generating reports that are closely aligned with the ground-truth report.

Additional qualitative analysis and visualizations are included in Figs.~\ref{fig:X_discussion} and \ref{fig:X_visualization} in the supplementary material.
\section{Conclusion}
We introduce a two-stage framework for radiology report generation, which combines disease-aware image-to-text retrieval with a self-correction module to refine generated reports by re-aligning reports with image features for greater accuracy and coherence. Our proposed framework achieves state-of-the-art performance on the MIMIC-CXR and IU X-ray benchmarks, generating clinically accurate, trustworthy  reports that can reduce radiologists’ workload. 

\section*{Acknowledgment}
This work was supported by the National Research Foundation of Korea (NRF) grants funded by the Korea government (MSIT) (Nos. RS-2023-00212498 and RS-2024-00415812), the Institute of Information \& Communications Technology Planning \& Evaluation (IITP) under Grant No. RS-2019-II190079, the Artificial Intelligence Graduate School Program at Korea University, and the MSIT (Ministry of Science and ICT), Korea, under the ITRC (Information Technology Research Center) support program (IITP-2024-RS-2024-00436857) supervised by the IITP.

\newpage

{
    \small
    \bibliographystyle{ieeenat_fullname}
    \bibliography{main}

\begin{thebibliography}{49}
\providecommand{\natexlab}[1]{#1}
\providecommand{\url}[1]{\texttt{#1}}
\expandafter\ifx\csname urlstyle\endcsname\relax
  \providecommand{\doi}[1]{doi: #1}\else
  \providecommand{\doi}{doi: \begingroup \urlstyle{rm}\Url}\fi

\bibitem[Achiam et~al.(2023)Achiam, Adler, Agarwal, Ahmad, Akkaya, Aleman, Almeida, Altenschmidt, Altman, Anadkat, et~al.]{achiam2023gpt}
Josh Achiam, Steven Adler, Sandhini Agarwal, Lama Ahmad, Ilge Akkaya, Florencia~Leoni Aleman, Diogo Almeida, Janko Altenschmidt, Sam Altman, Shyamal Anadkat, et~al.
\newblock {GPT}-4 technical report.
\newblock \emph{arXiv preprint arXiv:2303.08774}, 2023.

\bibitem[Anderson et~al.(2018)Anderson, He, Buehler, Teney, Johnson, Gould, and Zhang]{anderson2018bottom}
Peter Anderson, Xiaodong He, Chris Buehler, Damien Teney, Mark Johnson, Stephen Gould, and Lei Zhang.
\newblock Bottom-up and top-down attention for image captioning and visual question answering.
\newblock In \emph{Proceedings of the IEEE Conference on Computer Vision and Pattern Recognition}, pages 6077--6086, 2018.

\bibitem[Banerjee and Lavie(2005)]{banerjee2005meteor}
Satanjeev Banerjee and Alon Lavie.
\newblock {METEOR}: An automatic metric for {MT} evaluation with improved correlation with human judgments.
\newblock In \emph{Proceedings of the ACL Workshop on Intrinsic and Extrinsic Evaluation Measures for Machine Translation and/or Xummarization}, pages 65--72, 2005.

\bibitem[Chen et~al.(2020)Chen, Song, Chang, and Wan]{chen2020generating}
Zhihong Chen, Yan Song, Tsung-Hui Chang, and Xiang Wan.
\newblock Generating radiology reports via memory-driven transformer.
\newblock \emph{arXiv preprint arXiv:2010.16056}, 2020.

\bibitem[Chen et~al.(2022)Chen, Shen, Song, and Wan]{chen2022cross}
Zhihong Chen, Yaling Shen, Yan Song, and Xiang Wan.
\newblock Cross-modal memory networks for radiology report generation.
\newblock \emph{arXiv preprint arXiv:2204.13258}, 2022.

\bibitem[Cornia et~al.(2020)Cornia, Stefanini, Baraldi, and Cucchiara]{cornia2020meshed}
Marcella Cornia, Matteo Stefanini, Lorenzo Baraldi, and Rita Cucchiara.
\newblock Meshed-memory transformer for image captioning.
\newblock In \emph{Proceedings of the IEEE/CVF Conference on Computer Vision and Pattern Recognition}, pages 10578--10587, 2020.

\bibitem[Demner-Fushman et~al.(2016)Demner-Fushman, Kohli, Rosenman, Shooshan, Rodriguez, Antani, Thoma, and McDonald]{demner2016preparing}
Dina Demner-Fushman, Marc~D Kohli, Marc~B Rosenman, Sonya~E Shooshan, Laritza Rodriguez, Sameer Antani, George~R Thoma, and Clement~J McDonald.
\newblock Preparing a collection of radiology examinations for distribution and retrieval.
\newblock \emph{Journal of the American Medical Informatics Association}, 23\penalty0 (2):\penalty0 304--310, 2016.

\bibitem[Deng et~al.(2009)Deng, Dong, Socher, Li, Li, and Fei-Fei]{deng2009imagenet}
Jia Deng, Wei Dong, Richard Socher, Li-Jia Li, Kai Li, and Li Fei-Fei.
\newblock Image{N}et: A large-scale hierarchical image database.
\newblock In \emph{Proceedings of the IEEE Conference on Computer Vision and Pattern Recognition}, pages 248--255. IEEE, 2009.

\bibitem[Dosovitskiy et~al.(2020)Dosovitskiy, Beyer, Kolesnikov, Weissenborn, Zhai, Unterthiner, Dehghani, Minderer, Heigold, Gelly, et~al.]{alexey2020image}
Alexey Dosovitskiy, Lucas Beyer, Alexander Kolesnikov, Dirk Weissenborn, Xiaohua Zhai, Thomas Unterthiner, Mostafa Dehghani, Matthias Minderer, Georg Heigold, Sylvain Gelly, et~al.
\newblock An image is worth 16x16 words: Transformers for image recognition at scale.
\newblock \emph{arXiv preprint arXiv:2010.11929}, 2020.

\bibitem[He et~al.(2016)He, Zhang, Ren, and Sun]{he2016deep}
Kaiming He, Xiangyu Zhang, Shaoqing Ren, and Jian Sun.
\newblock Deep residual learning for image recognition.
\newblock In \emph{Proceedings of the IEEE Conference on Computer Vision and Pattern Recognition}, pages 770--778, 2016.

\bibitem[Hochreiter and Schmidhuber(1997)]{hochreiter1997long}
Sepp Hochreiter and J{\"u}rgen Schmidhuber.
\newblock Long short-term memory.
\newblock \emph{Neural Computation}, 9\penalty0 (8):\penalty0 1735--1780, 1997.

\bibitem[Hossain et~al.(2019)Hossain, Sohel, Shiratuddin, and Laga]{hossain2019comprehensive}
MD~Zakir Hossain, Ferdous Sohel, Mohd~Fairuz Shiratuddin, and Hamid Laga.
\newblock A comprehensive survey of deep learning for image captioning.
\newblock \emph{ACM Computing Surveys (CsUR)}, 51\penalty0 (6):\penalty0 1--36, 2019.

\bibitem[Hou et~al.(2024)Hou, Yan, Yan, Lang, and Zhou]{hou2024energy}
Zeyi Hou, Ruixin Yan, Ziye Yan, Ning Lang, and Xiuzhuang Zhou.
\newblock Energy-based controllable radiology report generation with medical knowledge.
\newblock In \emph{International Conference on Medical Image Computing and Computer-Assisted Intervention}, pages 240--250. Springer, 2024.

\bibitem[Irvin et~al.(2019)Irvin, Rajpurkar, Ko, Yu, Ciurea-Ilcus, Chute, Marklund, Haghgoo, Ball, Shpanskaya, et~al.]{irvin2019chexpert}
Jeremy Irvin, Pranav Rajpurkar, Michael Ko, Yifan Yu, Silviana Ciurea-Ilcus, Chris Chute, Henrik Marklund, Behzad Haghgoo, Robyn Ball, Katie Shpanskaya, et~al.
\newblock Che{XP}ert: A large chest radiograph dataset with uncertainty labels and expert comparison.
\newblock In \emph{Proceedings of the AAAI Conference on Artificial Intelligence}, pages 590--597, 2019.

\bibitem[Jing et~al.(2018)Jing, Xie, and Xing]{jing2017automatic}
Baoyu Jing, Pengtao Xie, and Eric Xing.
\newblock On the automatic generation of medical imaging reports.
\newblock In \emph{Proceedings of the 56th Annual Meeting of the Association for Computational Linguistics}, pages 2577--2586, 2018.

\bibitem[Johnson et~al.(2019)Johnson, Pollard, Greenbaum, Lungren, Deng, Peng, Lu, Mark, Berkowitz, and Horng]{johnson2019mimic}
Alistair~EW Johnson, Tom~J Pollard, Nathaniel~R Greenbaum, Matthew~P Lungren, Chih-ying Deng, Yifan Peng, Zhiyong Lu, Roger~G Mark, Seth~J Berkowitz, and Steven Horng.
\newblock {MIMIC-CXR-JPG}, a large publicly available database of labeled chest radiographs.
\newblock \emph{arXiv preprint arXiv:1901.07042}, 2019.

\bibitem[Li et~al.(2023{\natexlab{a}})Li, Lin, Chen, Lin, Liang, and Chang]{li2023dynamic}
Mingjie Li, Bingqian Lin, Zicong Chen, Haokun Lin, Xiaodan Liang, and Xiaojun Chang.
\newblock Dynamic graph enhanced contrastive learning for chest {X}-ray report generation.
\newblock In \emph{Proceedings of the IEEE/CVF Conference on Computer Vision and Pattern Recognition}, pages 3334--3343, 2023{\natexlab{a}}.

\bibitem[Li et~al.(2023{\natexlab{b}})Li, Liu, Wang, Chang, and Liang]{li2023auxiliary}
Mingjie Li, Rui Liu, Fuyu Wang, Xiaojun Chang, and Xiaodan Liang.
\newblock Auxiliary signal-guided knowledge encoder-decoder for medical report generation.
\newblock \emph{World Wide Web}, 26\penalty0 (1):\penalty0 253--270, 2023{\natexlab{b}}.

\bibitem[Lin(2004)]{lin2004rouge}
Chin-Yew Lin.
\newblock {ROUGE}: A package for automatic evaluation of summaries.
\newblock In \emph{Text Summarization Branches Out}, pages 74--81, 2004.

\bibitem[Liu et~al.(2024{\natexlab{a}})Liu, Tian, Chen, Song, and Zhang]{liu2024bootstrapping}
Chang Liu, Yuanhe Tian, Weidong Chen, Yan Song, and Yongdong Zhang.
\newblock Bootstrapping large language models for radiology report generation.
\newblock In \emph{Proceedings of the AAAI Conference on Artificial Intelligence}, pages 18635--18643, 2024{\natexlab{a}}.

\bibitem[Liu et~al.(2021{\natexlab{a}})Liu, Ge, and Wu]{liu2022competence}
Fenglin Liu, Shen Ge, and Xian Wu.
\newblock Competence-based multimodal curriculum learning for medical report generation.
\newblock In \emph{Proceedings of the 59th Annual Meeting of the Association for Computational Linguistics}, pages 3001--3012, 2021{\natexlab{a}}.

\bibitem[Liu et~al.(2021{\natexlab{b}})Liu, Wu, Ge, Fan, and Zou]{liu2021exploring}
Fenglin Liu, Xian Wu, Shen Ge, Wei Fan, and Yuexian Zou.
\newblock Exploring and distilling posterior and prior knowledge for radiology report generation.
\newblock In \emph{Proceedings of the IEEE/CVF Conference on Computer Vision and Pattern Recognition}, pages 13753--13762, 2021{\natexlab{b}}.

\bibitem[Liu et~al.(2021{\natexlab{c}})Liu, Yin, Wu, Ge, Zhang, and Sun]{liu2021contrastive}
Fenglin Liu, Changchang Yin, Xian Wu, Shen Ge, Ping Zhang, and Xu Sun.
\newblock Contrastive attention for automatic chest {X}-ray report generation.
\newblock In \emph{Findings of the Association for Computational Linguistics}, pages 269--280, 2021{\natexlab{c}}.

\bibitem[Liu et~al.(2024{\natexlab{b}})Liu, Li, Zhao, Chen, Chang, and Yao]{liu2024context}
Rui Liu, Mingjie Li, Shen Zhao, Ling Chen, Xiaojun Chang, and Lina Yao.
\newblock In-context learning for zero-shot medical report generation.
\newblock In \emph{Proceedings of the 32nd ACM International Conference on Multimedia}, pages 8721--8730, 2024{\natexlab{b}}.

\bibitem[Loshchilov and Hutter(2017)]{loshchilov2017decoupled}
Ilya Loshchilov and Frank Hutter.
\newblock Decoupled weight decay regularization.
\newblock \emph{arXiv preprint arXiv:1711.05101}, 2017.

\bibitem[Madaan et~al.(2024)Madaan, Tandon, Gupta, Hallinan, Gao, Wiegreffe, Alon, Dziri, Prabhumoye, Yang, et~al.]{madaan2024self}
Aman Madaan, Niket Tandon, Prakhar Gupta, Skyler Hallinan, Luyu Gao, Sarah Wiegreffe, Uri Alon, Nouha Dziri, Shrimai Prabhumoye, Yiming Yang, et~al.
\newblock Self-refine: Iterative refinement with self-feedback.
\newblock \emph{Advances in Neural Information Processing Systems}, 36, 2024.

\bibitem[Mao et~al.(2023)Mao, Mohri, and Zhong]{mao2023cross}
Anqi Mao, Mehryar Mohri, and Yutao Zhong.
\newblock Cross-entropy loss functions: Theoretical analysis and applications.
\newblock In \emph{International Conference on Machine Learning}, pages 23803--23828. PMLR, 2023.

\bibitem[Nguyen et~al.(2021)Nguyen, Nie, Badamdorj, Liu, Zhu, Truong, and Cheng]{nguyen2021automated}
Hoang Nguyen, Dong Nie, Taivanbat Badamdorj, Yujie Liu, Yingying Zhu, Jason Truong, and Li Cheng.
\newblock Automated generation of accurate {\&} fluent medical {X}-ray reports.
\newblock In \emph{Proceedings of the Conference on Empirical Methods in Natural Language Processing}, pages 3552--3569, 2021.

\bibitem[Pan et~al.(2020)Pan, Yao, Li, and Mei]{pan2020x}
Yingwei Pan, Ting Yao, Yehao Li, and Tao Mei.
\newblock X-linear attention networks for image captioning.
\newblock In \emph{Proceedings of the IEEE/CVF Conference on Computer Vision and Pattern Recognition}, pages 10971--10980, 2020.

\bibitem[Papineni et~al.(2002)Papineni, Roukos, Ward, and Zhu]{papineni2002bleu}
Kishore Papineni, Salim Roukos, Todd Ward, and Wei-Jing Zhu.
\newblock {BLEU}: a method for automatic evaluation of machine translation.
\newblock In \emph{Proceedings of the 40th annual meeting of the Association for Computational Linguistics}, pages 311--318, 2002.

\bibitem[Radford et~al.(2021)Radford, Kim, Hallacy, Ramesh, Goh, Agarwal, Sastry, Askell, Mishkin, Clark, et~al.]{radford2021learning}
Alec Radford, Jong~Wook Kim, Chris Hallacy, Aditya Ramesh, Gabriel Goh, Sandhini Agarwal, Girish Sastry, Amanda Askell, Pamela Mishkin, Jack Clark, et~al.
\newblock Learning transferable visual models from natural language supervision.
\newblock In \emph{International Conference on Machine Learning}, pages 8748--8763. PMLR, 2021.

\bibitem[Rennie et~al.(2017)Rennie, Marcheret, Mroueh, Ross, and Goel]{rennie2017self}
Steven~J Rennie, Etienne Marcheret, Youssef Mroueh, Jerret Ross, and Vaibhava Goel.
\newblock Self-critical sequence training for image captioning.
\newblock In \emph{Proceedings of the IEEE Conference on Computer Vision and Pattern Recognition}, pages 7008--7024, 2017.

\bibitem[Selvaraju et~al.(2017)Selvaraju, Cogswell, Das, Vedantam, Parikh, and Batra]{selvaraju2017grad}
Ramprasaath~R Selvaraju, Michael Cogswell, Abhishek Das, Ramakrishna Vedantam, Devi Parikh, and Dhruv Batra.
\newblock Grad-{CAM}: Visual explanations from deep networks via gradient-based localization.
\newblock In \emph{Proceedings of the IEEE International Conference on Machine Learning}, pages 618--626, 2017.

\bibitem[Shen et~al.(2024)Shen, Pei, Liu, and Tian]{shen2024automatic}
Hongyu Shen, Mingtao Pei, Juncai Liu, and Zhaoxing Tian.
\newblock Automatic radiology reports generation via memory alignment network.
\newblock In \emph{Proceedings of the AAAI Conference on Artificial Intelligence}, pages 4776--4783, 2024.

\bibitem[Tanida et~al.(2023)Tanida, M{\"u}ller, Kaissis, and Rueckert]{tanida2023interactive}
Tim Tanida, Philip M{\"u}ller, Georgios Kaissis, and Daniel Rueckert.
\newblock Interactive and explainable region-guided radiology report generation.
\newblock In \emph{Proceedings of the IEEE/CVF Conference on Computer Vision and Pattern Recognition}, pages 7433--7442, 2023.

\bibitem[Vaswani et~al.(2017)Vaswani, Shazeer, Parmar, Uszkoreit, Jones, Gomez, Kaiser, and Polosukhin]{vaswani2017attention}
Ashish Vaswani, Noam Shazeer, Niki Parmar, Jakob Uszkoreit, Llion Jones, Aidan~N Gomez, {\L}ukasz Kaiser, and Illia Polosukhin.
\newblock Attention is all you need.
\newblock \emph{Advances in Neural Information Processing Systems}, 30, 2017.

\bibitem[Vinyals et~al.(2015)Vinyals, Toshev, Bengio, and Erhan]{vinyals2015show}
Oriol Vinyals, Alexander Toshev, Samy Bengio, and Dumitru Erhan.
\newblock Show and tell: A neural image caption generator.
\newblock In \emph{Proceedings of the IEEE Conference on Computer Vision and Pattern Recognition}, pages 3156--3164, 2015.

\bibitem[Wang et~al.(2022)Wang, Tang, Wang, Li, and Zhou]{wang2022medical}
Zhanyu Wang, Mingkang Tang, Lei Wang, Xiu Li, and Luping Zhou.
\newblock A medical semantic-assisted transformer for radiographic report generation.
\newblock In \emph{International Conference on Medical Image Computing and Computer-Assisted Intervention}, pages 655--664. Springer, 2022.

\bibitem[Wang et~al.(2023)Wang, Liu, Wang, and Zhou]{wang2023metransformer}
Zhanyu Wang, Lingqiao Liu, Lei Wang, and Luping Zhou.
\newblock {MET}ransformer: Radiology report generation by transformer with multiple learnable expert tokens.
\newblock In \emph{Proceedings of the IEEE/CVF Conference on Computer Vision and Pattern Recognition}, pages 11558--11567, 2023.

\bibitem[Welleck et~al.(2022)Welleck, Lu, West, Brahman, Shen, Khashabi, and Choi]{welleck2022generating}
Sean Welleck, Ximing Lu, Peter West, Faeze Brahman, Tianxiao Shen, Daniel Khashabi, and Yejin Choi.
\newblock Generating sequences by learning to self-correct.
\newblock \emph{arXiv preprint arXiv:2211.00053}, 2022.

\bibitem[Xu et~al.(2015)Xu, Ba, Kiros, Cho, Courville, Salakhutdinov, Zemel, and Bengio]{xu2015show}
Kelvin Xu, Jimmy Ba, Ryan Kiros, Kyunghyun Cho, Aaron~C. Courville, Ruslan Salakhutdinov, Richard~S. Zemel, and Yoshua Bengio.
\newblock Show, attend and tell: Neural image caption generation with visual attention.
\newblock In \emph{Proceedings of the IEEE International Conference on Machine Learning}, 2015.

\bibitem[Xue and Huang(2019)]{xue2019improved}
Yuan Xue and Xiaolei Huang.
\newblock Improved disease classification in chest {X}-rays with transferred features from report generation.
\newblock In \emph{Information Processing in Medical Imaging}, pages 125--138. Springer, 2019.

\bibitem[Xue et~al.(2018)Xue, Xu, Rodney~Long, Xue, Antani, Thoma, and Huang]{xue2018multimodal}
Yuan Xue, Tao Xu, L Rodney~Long, Zhiyun Xue, Sameer Antani, George~R Thoma, and Xiaolei Huang.
\newblock Multimodal recurrent model with attention for automated radiology report generation.
\newblock In \emph{International Conference on Medical Image Computing and Computer-Assisted Intervention}, pages 457--466. Springer, 2018.

\bibitem[Yan et~al.(2024)Yan, Gu, Zhu, and Ling]{yan2024corrective}
Shi-Qi Yan, Jia-Chen Gu, Yun Zhu, and Zhen-Hua Ling.
\newblock Corrective retrieval augmented generation.
\newblock \emph{arXiv preprint arXiv:2401.15884}, 2024.

\bibitem[Yang et~al.(2023)Yang, Wu, Ge, Zheng, Zhou, and Xiao]{yang2023radiology}
Shuxin Yang, Xian Wu, Shen Ge, Zhuozhao Zheng, S~Kevin Zhou, and Li Xiao.
\newblock Radiology report generation with a learned knowledge base and multi-modal alignment.
\newblock \emph{Medical Image Analysis}, 86:\penalty0 102798, 2023.

\bibitem[Yang et~al.(2016)Yang, Yuan, Wu, Cohen, and Salakhutdinov]{yang2016review}
Zhilin Yang, Ye Yuan, Yuexin Wu, William~W Cohen, and Russ~R Salakhutdinov.
\newblock Review networks for caption generation.
\newblock \emph{Advances in Neural Information Processing Systems}, 29, 2016.

\bibitem[Yin et~al.(2019)Yin, Qian, Wei, Li, Zhang, Li, and Zheng]{yin2019automatic}
Changchang Yin, Buyue Qian, Jishang Wei, Xiaoyu Li, Xianli Zhang, Yang Li, and Qinghua Zheng.
\newblock Automatic generation of medical imaging diagnostic report with hierarchical recurrent neural network.
\newblock In \emph{Proceedings of the IEEE International Conference on Data Mining (ICDM)}, pages 728--737. IEEE, 2019.

\bibitem[You et~al.(2021)You, Liu, Ge, Xie, Zhang, and Wu]{you2021aligntransformer}
Di You, Fenglin Liu, Shen Ge, Xiaoxia Xie, Jing Zhang, and Xian Wu.
\newblock Align{T}ransformer: Hierarchical alignment of visual regions and disease tags for medical report generation.
\newblock In \emph{International Conference on Medical Image Computing and Computer-Assisted Intervention}, pages 72--82. Springer, 2021.

\bibitem[Zhang et~al.(2020)Zhang, Wang, Xu, Yu, Yuille, and Xu]{zhang2020radiology}
Yixiao Zhang, Xiaosong Wang, Ziyue Xu, Qihang Yu, Alan Yuille, and Daguang Xu.
\newblock When radiology report generation meets knowledge graph.
\newblock In \emph{Proceedings of the AAAI Conference on Artificial Intelligence}, pages 12910--12917, 2020.

\end{thebibliography}
}

\clearpage
\appendix

\newpage
\setcounter{page}{1}
\maketitlesupplementary

\section{Contrastive Loss}
The contrastive loss is based on the CLIP loss~\cite{radford2021learning}, which maximizes the cosine similarity between paired image-text features (positive pairs, i.e., an image and its corresponding report) while minimizing the similarity between unpaired image-text features. The contrastive loss $\mathcal{L}_{\text{con}}$ can be expressed as:

\begin{equation}
\begin{split}
\mathcal{L}_{\text{con}} = -\frac{1}{2} ( \log \frac{e^{(\textit{sim}(\mathbf{f}_I, \mathbf{f}_T)/\tau)}}{\sum_{j=1}^{q} e^{(\textit{sim}(\mathbf{f}_I, \mathbf{f}_T^j)/\tau)}} \\
 + \log \frac{e^{(\textit{sim}(\mathbf{f}_I, \mathbf{f}_T)/\tau)}}{\sum_{j=1}^{q}   e^{(\textit{sim}(\mathbf{f}_I^j, \mathbf{f}_T)/\tau)}} ),
\end{split}
\end{equation}
where $\tau$ is a learnable temperature parameter, $\mathbf{f}_I$ and $\mathbf{f}_T$ are image and text features from the input image and its corresponding report, $\mathbf{f}^j_I$ and $\mathbf{f}^j_T$ are the $j^{th}$ image and text features stored in the training queue, $q$ is the number of features in the queue, and $\textit{sim}$ represents the cosine similarity between two features. The cosine similarity between features from the input image and its corresponding report is defined as:

\begin{equation}
\textit{sim}(\mathbf{f}_I, \mathbf{f}_T) = \frac{\mathbf{f}_I \cdot \mathbf{f}_T}{|\mathbf{f}_I| \cdot |\mathbf{f}_T|}.
\end{equation}

\section{Generation Loss}

We employ a cross-entropy loss, denoted as $\mathcal{L}_{gen}$, to train the text generator for synthesizing accurate and trustworthy radiology reports. This loss minimizes the discrepancy between the generated report $\hat{T}$ and the ground-truth report $T$, which consists of $l$ tokens $T = \{T_1, T_2, ..., T_l\}$. At each time step $t$, the model predicts the probability of the next token $T_t$ conditioned on all previous tokens $T_1, T_2, ..., T_{t-1}$. The generation loss can be defined as:

\begin{equation}
\mathcal{L}_{gen} = - \sum_{t=1}^{l} \log P(T_t \mid T_1,...,T_{t-1}, \mathbf{f}_D, \mathbf{f}_T, \mathbf{f}^1_{\hat{T}},...,\mathbf{f}^k_{\hat{T}}),
\end{equation}
where $T_t$ is the $t^{th}$ token in the ground-truth report $T$, $T_1, …, T_{t-1}$ represent all preceding tokens, $\mathbf{f}_D$ represents the disease-relevant features, $\mathbf{f}_T$ denotes the text features, $\mathbf{f}^1_{\hat{T}}, …, \mathbf{f}^k_{\hat{T}}$ are the retrieved text features, and $l$ is the length of the ground-truth report.

\section{Qualitative Analysis}
\begin{figure*}[p]
	\begin{center}
		\includegraphics[width=0.8\linewidth]{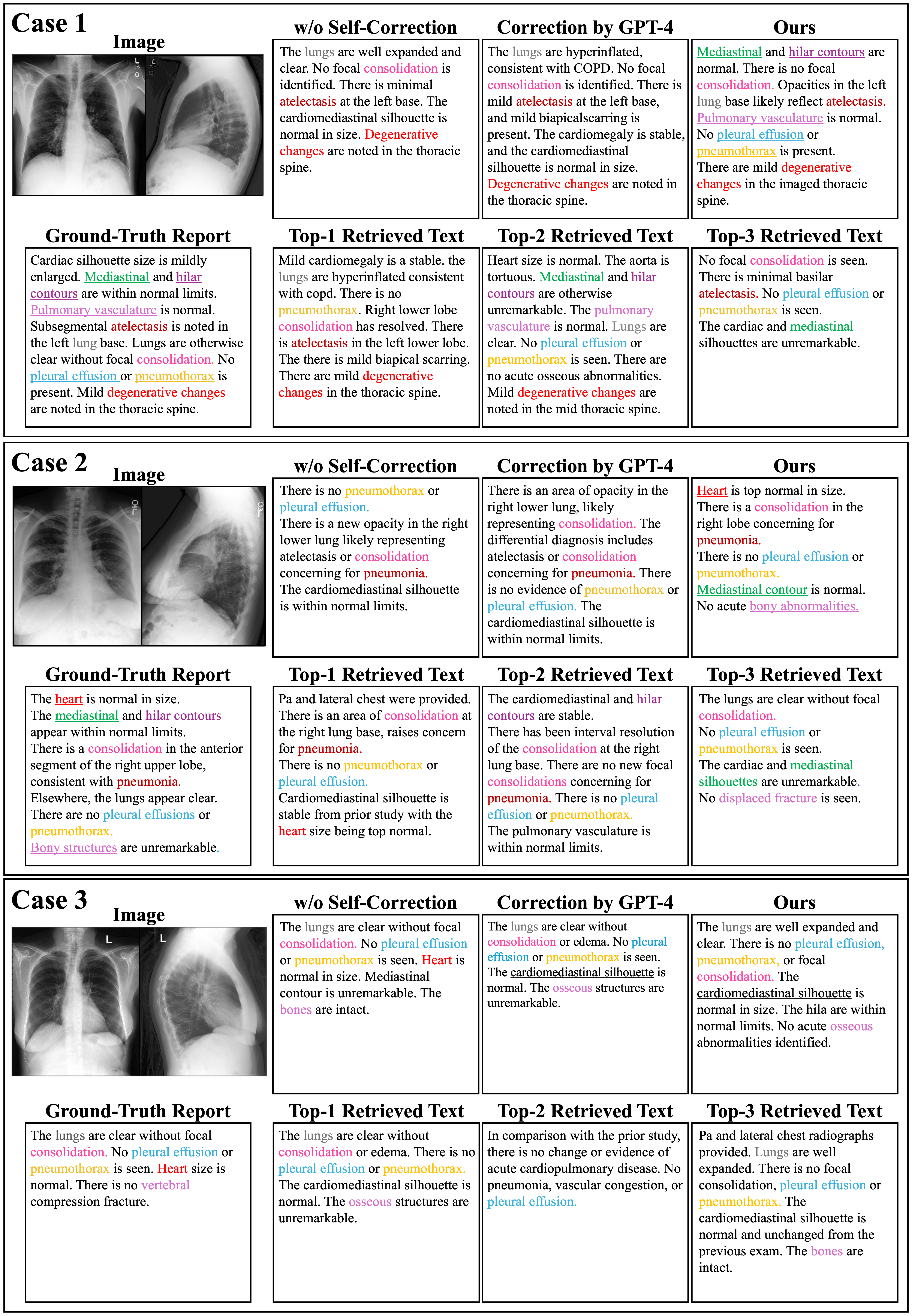}
	\end{center}
	\caption{An additional qualitative analysis of reports for three samples from the MIMIC-CXR dataset is presented. The top row of each sample displays an image set from two different views alongside a generated report from our proposed framework without the self-correction module (``w/o Self-Correction”). We further attempted to refine the generated report of ``w/o Self-Correction”  using GPT-4 \cite{achiam2023gpt} (``Correction by GPT-4") to compare it with the generated report from our proposed framework with self-correction (``Ours”). The bottom row shows the ground-truth report and the Top-3 retrieved texts from image-to-text retrieval. Key findings are highlighted in different colors for clarity.}
	\label{fig:X_discussion}
\end{figure*}

Fig. \ref{fig:X_discussion} presents an additional qualitative analysis of generated reports of three cases from the MIMIC-CXR dataset, including generated report from our proposed framework without self-correction (``w/o Self-Correction''). We refine the generated report of ``w/o Self-Correction" by GPT-4 \cite{achiam2023gpt} (``Correction by GPT-4''), and our proposed framework with self-correction (``Ours''). We also show the ground-truth report and the Top-3 retrieved reports from image-to-text retrieval.

\noindent\textbf{Details for Correction by GPT-4}
We evaluate the refinement of generated reports using GPT-4 \cite{achiam2023gpt}. The goal is to assess whether large language models (LLMs) can effectively improve the quality of the generated reports by addressing omissions and enhancing coherence. We provide GPT-4 with the generated report, retrieved texts, and the input image, using the following structured prompt:

\begin{quote}
\textit{[the input image] Retrieved Patient’s Text Top-1: [the retrieved text (top-1)]. ... Retrieved Patient’s Text Top-k: [the retrieved text (top-k)]. If the generated report is [the generated report], correct the generated report.}
\end{quote}

Here, the prompt includes the input image, the top-\(k\) retrieved texts from image-to-text retrieval, which provide contextual information relevant to the input image, and the generated report from our proposed framework without self-correction (``w/o Self-Correction'').

\begin{figure*}[t]
	\begin{center}
		\includegraphics[width=\linewidth]{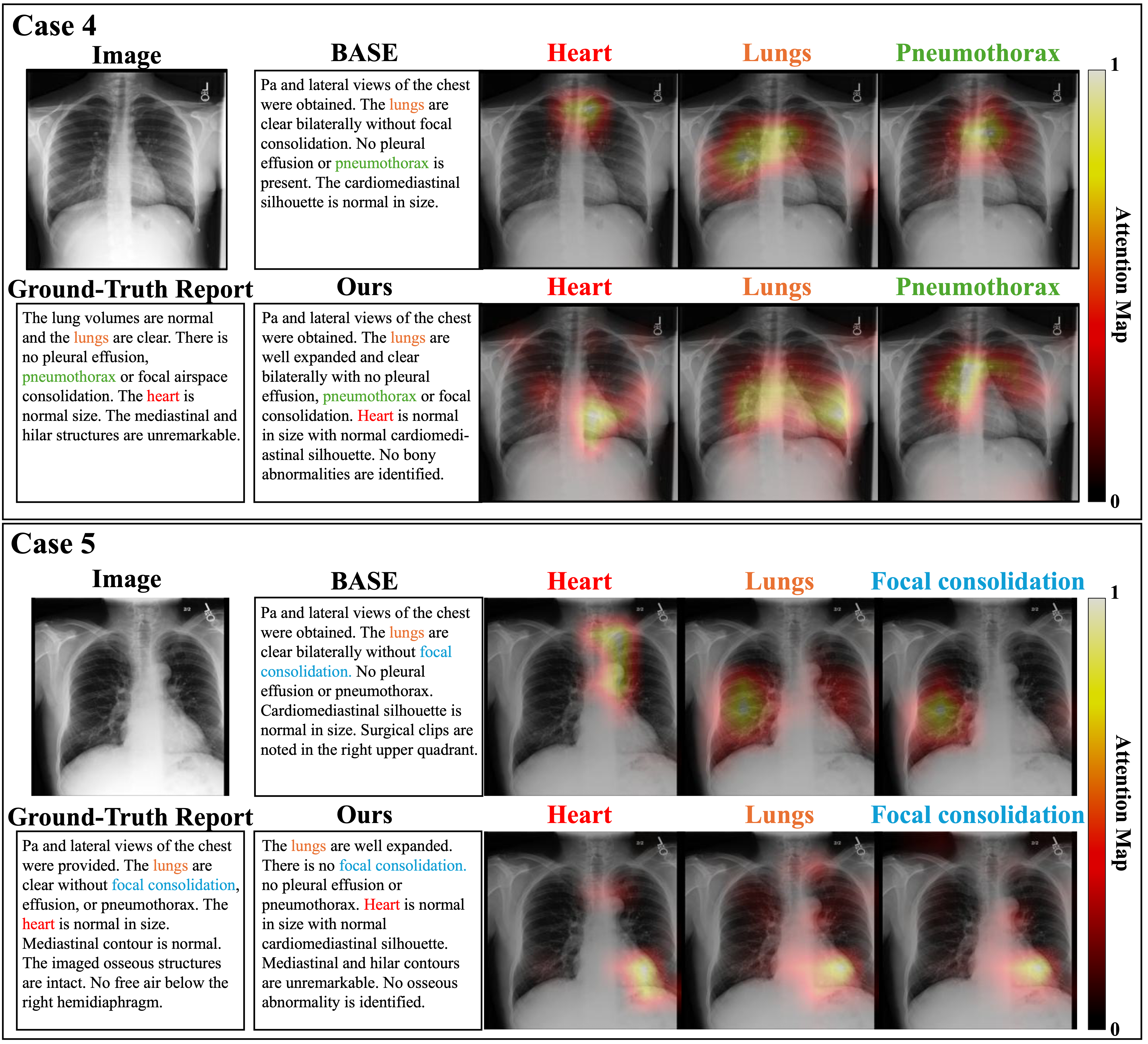}
	\end{center}
	\caption{Visualizations of the generated reports and attention maps from the baseline model (BASE) and our proposed framework (Ours) on two samples from the MIMIC-CXR dataset. The attention maps, visualized using Grad-CAM \cite{selvaraju2017grad}, illustrate the regions that BASE and Ours focuses on according to keywords such as ``heart,” ``lung,” ``pneumothorax,” and ``focal consolidation,” with each keyword highlighted in different colors.}
	\label{fig:X_visualization}
\end{figure*}

\noindent\textbf{Case 1}
The ``w/o Self-Correction'' report provides a basic assessment, accurately identifying key findings such as ``lungs'' and `` atelectasis at the left base.'' However, it omits details regarding ``pulmonary vasculature,'' ``pleural effusion'' and ``pneumothorax,'' which are critical for specific analysis. On the other hand, ``Correction by GPT-4'' introduces additional observations, such as ``hyperinflated, consistent with COPD'' and ``mild biapical scarring,'' which are not consistent with the ground-truth.

In contrast, ``Ours'' generates a report that aligns with the ground-truth and accurately captures key findings. It not only confirms the absence of ``pleural effusion'' and ``pneumothorax'' but also identifies detail observations such as ``opacities in the left lung base likely reflect atelectasis'' and ``pulmonary vasculature is normal,'' which are consistent with the ground-truth.  Additionally, ``Ours'' accurately captures the description of ``mediastinal and hilar contours are normal,'' demonstrating its ability to comprehensively address key disease-relevant findings, further enhancing its alignment with the ground-truth.

In both ``w/o Self-Correction'' and ``Ours,'' the Top-3 retrieved reports provide additional contextual information and contain key findings aligned with the ground-truth report, such as ``Degenerative changes'' and ``atelectasis.'' This also demonstrates that our proposed framework effectively leverages the retrieved reports similar to the ground-truth.

\noindent\textbf{Case 2}
The ``w/o Self-Correction'' report identifies essential findings such as the absence of ``pneumothorax and pleural effusion.'' However, it does not comprehensively address ``mediastinal contour'' or ``bony structures.'' Similarly, ``Correction by GPT-4'' refines the phrasing of findings, such as describing the opacity as ``likely representing consolidation.'' However, it produces redundancy and does not explicitly describe some key findings, such as ``mediastinal contour and the ``bony structures.''

In contrast, ``Ours'' generates a report that aligns with the ground-truth and accurately captures the patient's condition. It not only identifies the absence of ``pleural effusion and pneumothorax,'' but also describes the ``mediastinal contour'' as normal and uniquely includes a statement about the absence of acute ``bony abnormalities,'' aligning with the ground-truth, such as ``bony structures are unremarkable.'' 

In both ``w/o Self-Correction'' and ``Ours,'' the Top-3 retrieved reports provide additional contextual information and contain key findings aligned with the ground-truth report, such as ``consolidation,'' ``pneumothorax,'' ``pleural effusion,'' and ``pneumonia.'' This also demonstrates that our proposed framework effectively leverages the retrieved reports, which are similar to the ground-truth.

\noindent\textbf{Case 3}
``w/o Self-Correction'' successfully captures key findings from the ground-truth, such as ``pleural effusion,'' ``pneumothorax,'' and ``consolidation.'' However, both ``Correction by GPT-4'' and ``Ours'' generate the phrase ``cardiomediastinal silhouette'' instead of ``heart.'' Similarly, while the retrieved texts effectively capture key findings from the ground-truth report, such as ``pleural effusion'' and ``pneumothorax,'' they include ``cardiomediastinal silhouette'' instead of ``heart.'' The term ``cardiomediastinal silhouette'' can be used as an indirect indicator for assessing ``heart size.'' Since the retrieved texts do not include the direct keyword ``heart,'' self-correction mechanisms, both ``Correction by GPT-4'' and ``Ours,'' generate an indirect term instead.

This case highlights the importance of designing self-correction mechanisms to prioritize the retrieval of reports that explicitly include key findings from the ground-truth. Accurate retrieval is crucial for ensuring that generated reports align closely with disease-relevant findings. While our proposed framework demonstrates significant improvements in capturing these findings, this example underscores the need to refine the retrieval to directly align with the ground-truth report in the self-correction process.

\section{Attention Visualization}
Fig. \ref{fig:X_visualization} presents an additional attention visualization using Grad-CAM~\cite{selvaraju2017grad} to compare the BASE setting (``BASE'') and our proposed framework (``Ours'') for radiology report generation. BASE setting includes only the classification loss and generation loss. The visualization highlights the regions of focus for three critical keywords with each keyword represented in a distinct color for clarity.

\noindent\textbf{Case 4}
Both models successfully generate the keywords ``lungs'' and ``pneumothorax,'' aligning with the ground-truth report. However, the baseline model misses ``heart,'' while our proposed model accurately captures it. This difference is reflected in the attention maps: our proposed model focuses on the actual heart region, as well as ``lungs'' and ``pneumothorax,'' whereas the baseline model fails to attend to the heart region. These results demonstrate the effectiveness of our proposed model in capturing disease-related findings.

\noindent\textbf{Case 5}
Both ``BASE'' and ``Ours'' successfully generate the keywords ``lungs'' and ``focal consolidation,'' aligning with the ground-truth report. However, the attention maps again highlight notable differences. Similar to Case 4, the ``BASE'' model attends predominantly to regions associated with the ``lungs'' but fails to focus on key areas related to the ``heart.'' Additionally, its attention for ``focal consolidation'' is similar with the regions of  ``lungs.''

For ``Ours,'' the attention maps exhibit strong focus on the ``heart,'' demonstrating the ability of our proposed framework to identify and prioritize critical regions for this keyword. However, for ``lungs'' and ``focal consolidation,'' the attention maps show some focus on irrelevant regions. Despite this limitation, our proposed framework successfully generates the keywords ``lungs'' and ``focal consolidation,'' which are clinically accurate and align with the ground-truth report. This highlights the inherent difficulty of extracting disease-relevant features directly from X-ray images. It also highlights the effectiveness of our proposed framework compared to ``BASE,’’ particularly in leveraging retrieved reports and self-correction mechanisms to supplement and guide the report generation process, thereby compensating for potential inconsistencies with image features.

\begin{table*}[]
\centering
\begin{tabular}{c|c|cccc|cccccc}
\hline
Dataset                   & Setting & CL         & I2T        & DM         & SC         & BLEU-1         & BLEU-2         & BLEU-3         & BLEU-4         & RG-L           & METEOR         \\ \hline
\multirow{5}{*}{IU X-ray} & BASE    & -          & -          & -          & -          & 0.421          & 0.271          & 0.183          & 0.124          & 0.326          & 0.169          \\
                          & (a)     & \checkmark & -          & -          & -          & 0.427          & 0.282          & 0.195          & 0.137          & 0.355          & 0.169          \\
                          & (b)     & \checkmark & \checkmark & -          & -          & 0.464          & 0.320          & 0.230          & 0.174          & 0.358          & 0.185          \\
                          & (c)     & \checkmark & \checkmark & \checkmark & -          & 0.472          & 0.328          & 0.240          & 0.182          & 0.386          & 0.201          \\ \cline{2-12} 
                          & (d)     & \checkmark & \checkmark & \checkmark & \checkmark & \textbf{0.486} & \textbf{0.348} & \textbf{0.265} & \textbf{0.208} & \textbf{0.411} & \textbf{0.205} \\ \hline
\end{tabular}
\caption{An ablation study of our proposed framework on the IU X-ray dataset, assessing the impact of key components: contrastive loss (CL), image-to-text retrieval (I2T), disease-matching constraint (DM), and self-correction (SC). A ``\checkmark" indicates the presence of each component, while ``-" denotes its absence. The BASE setting involves training only with the classification loss and the generation loss.}
\label{tab:X_table}
\end{table*}

\section{Ablation Study on IU X-ray}

We extend our ablation study to the IU X-ray dataset to evaluate the incremental impact of each component in our proposed framework: contrastive loss (CL), image-to-text retrieval (I2T), disease-matching constraint (DM), and self-correction (SC). The results are summarized in Table \ref{tab:ablation}, showing performance improvements as these components are progressively added to the BASE setting, which includes only the classification loss and generation loss.

Starting from the BASE setting, which achieves BLEU-4 of 0.124 and ROUGE-L of 0.326, the addition of contrastive learning (CL) in setting (a) leads to modest improvements in BLEU-4 (0.137) and ROUGE-L (0.355). This indicates that aligning image and text embeddings through contrastive learning enhances feature representation, which aids the downstream generation task.

Adding image-to-text retrieval (I2T) in setting (b) significantly boosts performance across all metrics, with BLEU-4 increasing to 0.174 and ROUGE-L to 0.358. This demonstrates the value of retrieving disease-relevant reports, which provide additional contextual information for accurate report generation.

In setting (c), the inclusion of the disease-matching constraint (DM) further improves performance, with BLEU-4 reaching 0.182 and ROUGE-L increasing to 0.386. The disease-matching constraint ensures that the retrieved reports align more closely with the disease-relevant findings of the input images, resulting in more accurate and clinically coherent generated reports.

Finally, adding self-correction (SC) in setting (d) achieves the best results, with BLEU-4 improving to 0.208 and ROUGE-L reaching 0.411. This substantial improvement highlights the effectiveness of the self-correction module in refining the generated reports. By re-aligning the generated reports with the input image features in the embedding space, the self-correction module reduces discrepancies and enhances the accuracy and coherence of the generated reports.

This ablation study on the IU X-ray dataset demonstrates the consistent effectiveness of each component in our proposed framework. In other words, this study validates the importance of integrating contrastive learning, disease-aware retrieval, disease-matching, and self-correction to achieve state-of-the-art performance in radiology report generation.

\begin{figure}[!t]
	\begin{center}
		\includegraphics[width=\linewidth]{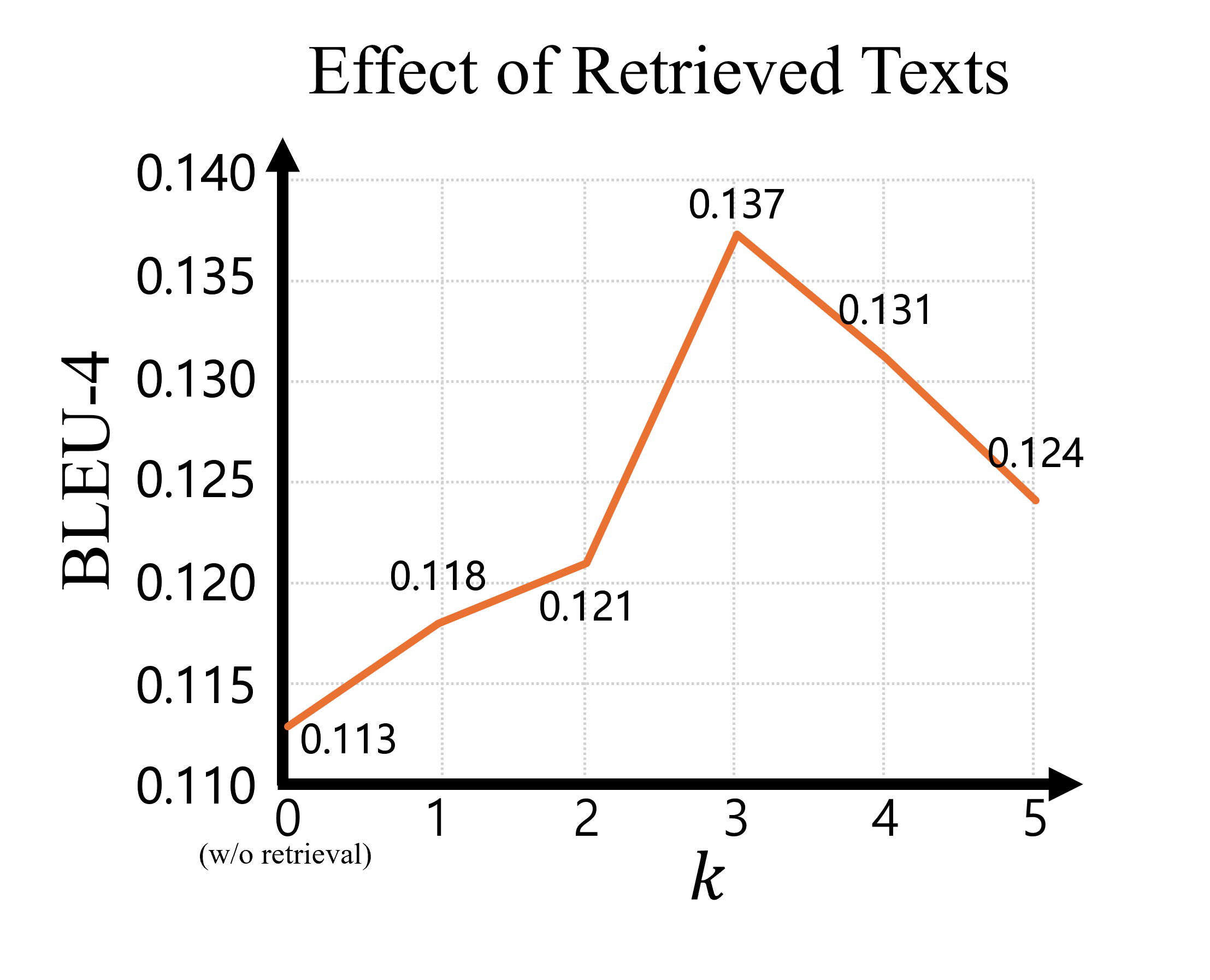}
	\end{center}
	\caption{We evaluate the effect of the number of retrieved texts ($k$) on BLEU-4 performance for the MIMIC-CXR dataset in our proposed framework.}
	\label{fig:X_k}
\end{figure}

\section{Effect of Retrieved Texts}

Our proposed framework retrieves similar texts based on input images to generate accurate reports. Fig. \ref{fig:X_k} evaluates the effect of retrieved texts, ranging from \(k=0\) (without retrieval) to \(k=5\), on the BLEU-4 performance. It demonstrates that retrieving texts (\(k=1,2,..,5\)) enhances the BLEU-4 score compared to the performance without retrieval (\(k=0\)).

In detail, the BLEU-4 score for \(k=0\)  (without retrieval) is 0.113, which is significantly lower than the BLEU-4 scores achieved when retrieval is employed. This underscores the importance of retrieval in our proposed framework. The retrieved texts provide critical disease-relevant findings that enhance the alignment between the generated reports and the ground-truth findings, thereby improving performance for report generation.

The BLEU-4 score gradually increases as \( k \) increases from 1 to 3, suggesting that retrieving more texts provides additional useful context for generating accurate radiology reports. However, when \( k \) exceeds 3, a decline in performance is observed. Our possible explanation is that the additional retrieved texts beyond \( k \) = 3 may include less relevant information, which could dilute the effectiveness of disease-relevant findings.

In summary, this analysis highlights the importance of the retrieval process in providing relevant textual information and demonstrates its crucial role in generating accurate and comprehensive radiology reports.
\end{document}